\documentclass[a4paper]{article}
\usepackage{comment}
\usepackage{INTERSPEECH2021}

\title{Exploring the Potential of Lexical Paraphrases for \\ Mitigating Noise-Induced Comprehension Errors}
\name{Anupama Chingacham, Vera Demberg, Dietrich Klakow }
\address{
  Saarland Informatics Campus, Saarland University, Germany}
\email{achingacham@lsv.uni-saarland.de}

\begin{document}

\maketitle
\begin{abstract}

Listening in noisy environments can be difficult even for individuals with a normal hearing thresholds. The speech signal can be masked by noise, which may lead to word misperceptions on the side of the listener, and overall difficulty to understand the message. To mitigate hearing difficulties on listeners, a co-operative speaker utilizes voice modulation strategies like Lombard speech to generate noise-robust utterances, and similar solutions have been developed for speech synthesis systems. In this work, we propose an alternate solution of choosing noise-robust lexical paraphrases to represent an intended meaning. Our results show that lexical paraphrases differ in their intelligibility in noise. We evaluate the intelligibility of synonyms in context and find that choosing a lexical unit that is less risky to be misheard than its synonym introduced an average gain in comprehension of 37\% at SNR~-5 dB and 21\% at SNR~0 dB for babble noise. 

\end{abstract}

\section{Introduction}

In the event of speech distortions caused by echoes, reverberations, or unavoidable background noise like in a busy cafeteria or traffic, speech perception can be challenging even for individuals with normal hearing thresholds. When noise hinders listening, the meaning of a message perceived by a listener can be different from the actual meaning intended by the speaker, and hence lead to misunderstandings or even conversation breakdowns in extreme cases\cite{grimshaw1980mishearings}. It is important to take measures that reduce the adverse effects of noise in conversations.

In comparison to listeners, speakers in noise environments have better control over those noise-induced comprehension errors, as they can modify an utterance to be noise-robust. Previous studies on speech production in noise have shown that Lombard speech and its effects can improve intelligibility in noise with the modification of acoustic features such as pitch, loudness, and duration \cite{brumm2011evolution, zollinger2011lombard}. However, studies have also showcased that Lombard benefits vary with linguistic content \cite{patel2008influence, valentini2014using}.
A second factor that can affect word recognition in noise is linguistic characteristics such as predictability \cite{kalikow1977development}, syntactic complexity \cite{uslar2011does, carroll2013effects, van2018speech}, phonetic and lexical variables such as frequency and neighborhood density \cite{luce1998recognizing, mcardle2008predicting}. 
While a lot of work in speech synthesis has focused on the first aspect, we here propose to exploit the second factor: \textit{we check whether a more easily intelligible alternate word (a synonym) is available, and generate utterances after replacing such words by their more noise-robust synonym}. 

Earlier work in speech perception in noise has investigated how different  speech tokens such as vowels \cite{doi:10.1121/1.1908983,doi:10.1121/1.1810292} or consonants \cite{weber2003consonant,doi:10.1121/1.3224721} are affected by noise, and have considered word intelligibility in isolation \cite{luce1998recognizing, lexicalneighborhoods,wilson2008comparison} as well as in context \cite{kalikow1977development}. 
Although earlier studies on word misperception in noise \cite{doi:10.1121/1.4809540, doi:10.1121/1.5090196, marxer2016corpus, cooke2009discovering} have shown that the noise impact is dependent on the lexical items, to the best of our knowledge, the current work constitutes the first study to explore the potential of lexical paraphrases to reduce noise-induced comprehension errors.

In this paper, the effect of lexical choice on noise-induced comprehension errors is tested by comparing the intelligibility of synonyms (in English), as they support a simple method of substituting linguistic form while keeping the intended meaning of an utterance. 
To this end, listening experiments 
\footnote{Listening experiment data can be found under:
\\
https://tinyurl.com/4tm9bwbf } 
with human subjects were conducted at 3 different levels of babble noise: (i) high (SNR~-5), (ii) medium (SNR~0), and (iii) low (SNR~5) / no noise. 
Since it is known that predictability can influence word recognition, we investigated the recognition difference between synonyms in noise by conducting two sets of listening experiment with synonyms; (i) without context, where we demonstrate that synonyms indeed differ in their intelligibility in noise (see Section \ref{sec:wo_context}) and (ii)  with context, where we aim to show that synonym replacements can improve intelligibility in naturalistic contexts (reported in Section \ref{sec:with_context}). Similar to prior work on ``slips of the ear'' \cite{vitevitch2002naturalistic}, as a measure of intelligibility in the experiments, we calculate the proportion of the number of correct identifications of a word compared to all instances of that word in a noisy environment. We refer this measure as Human Recognition Score (HRS).


In order to use synonym replacement in practical applications, it is necessary to be able to predict the intelligibility of a word in a specific noisy environment. Section \ref{sec:modeling} describes our experiment with computational measures designed to study whether the lexical, linguistic and acoustic features of a word equally influences its recognition in different noisy environments. 
Additionally, we examined the significance of each of these features in contributing to the improved comprehension by lexical replacements in noise.

\section{Synonyms \textit{without} context}
\label{sec:wo_context}
In order for synonym replacement to be a promising approach, we need to test whether synonyms substantially differ with respect to their intelligibility in noise. Only if this is the case, it does make sense to attempt to replace one for the other. 
In our experiment, synonyms were presented separately to different participants as spoken words in three different listening environments; clean (no noise), babble noise at SNR~0 and SNR~-5.

\textbf{Stimuli:} Lexical items for this experiment were generated by selecting the most frequent words in a spoken corpus, Verbmobil \cite{verbmobil}. In order to make sure that these words can later be substituted reliably without changing the meaning of utterances, we further selected only those words that belong to a single synset in the lexical database WordNet \cite{wordnet}. 

A set of 189 synonym pairs (265 unique words) were selected and split into multiple lists such that no two synonyms were presented to the same participants.
Stimuli for this experiment consisted of spoken words which were synthesized using the Google Translate API (gTTS) \cite{gTTS} and their noisy signals generated by performing additive noise mixing with babble noise from NOISEX-92 database \cite{varga1993assessment} . 

\textbf{Design and procedure:} These speech signals were categorized into multiple blocks and it was ensured each block was presented to five different listeners.
Participants were instructed to write down what they understood after listening to each spoken word. In the study instructions, we asked participants to ensure a quiet environment and to use a good quality headphone in order to take part in the experiment. Also, the significance of these recommendations were highlighted  to them by providing sample audio files and a warning that audio files will be played only once.

\textbf{Participants:} The single word listening experiment was conducted on the crowd-sourcing platform Prolific \cite{palan2018prolific} with 75 native British English speakers (53 females and 22 males) with an average age of 31 (ranges from 18 to 49).

\textbf{Analysis:} By comparing the phonetic transcription of a target word $i$ and its 5 responses, the Human Recognition Score (HRS) for each stimulus (\textit{i}) was calculated as defined in (\ref{eq:HRS}).

\begin{equation}
\label{eq:HRS}
    \textrm{HRS}_{i} = \frac{\textrm{\# correct.responses}_{i}}{\textrm{\# total.responses}_{i}}
\end{equation}

For each pair of synonyms, we calculated the difference in their HRS, to quantify how much these more intelligible synonym differed from the harder to understand variant. Hence, this value ranges from 0.0 to 1.0, representing \textit{no gain} to \textit{maximum gain} in comprehension by choosing a lexical unit over its less recognized synonym. 

\begin{equation}
\label{eq:diff.HRS}
    \textrm{diff.HRS}_{(s_{1}, s_{2})} =
    abs( \textrm{HRS}_{s_{1}} - \textrm{HRS}_{s_{2}} )
\end{equation}

\subsection{Results}

\begin{figure}
    \centering
    \includegraphics[width=0.75\linewidth]{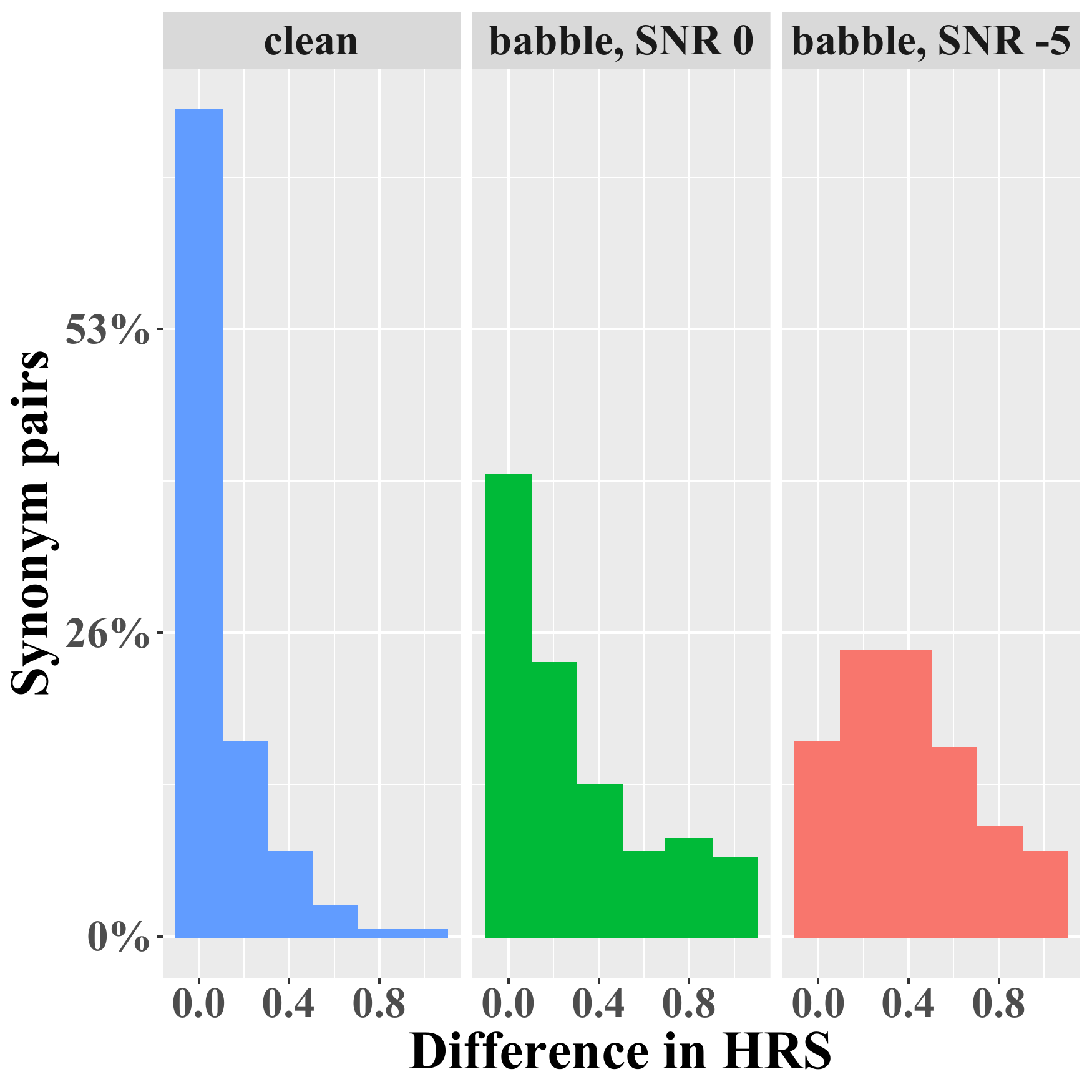}
    \caption{The number of synonym pairs that were distinct in recognition significantly increased with an increase in babble noise level (clean $\longrightarrow$ SNR 0 $\longrightarrow$ SNR -5), when they were presented \textbf{without context} in different listening environments}
    \label{fig:SWL_diffHRS_histogram}
\end{figure}

With the increase in noise (clean $\longrightarrow$ SNR~0 $\longrightarrow$ SNR~-5), as expected, the average HRS reduced significantly from 0.93 to 0.83 and 0.57 (\textit{p} $<$ 0.001), as the increased masking effect of noise tampered word recognition.
The average recognition difference between synonym pairs steadily increased from 0.09 to 0.28 (\textit{p} $<$ 0.05) and finally to 0.39 (\textit{p} $<$ 0.05). 
As presented in Figure~\ref{fig:SWL_diffHRS_histogram}, in both noisy environments, there is a significant increase in the number of synonym pairs which were distinct in their HRSs than those in clean.

The increased number of synonyms which differ substantially in intelligibility at SNR~0 and SNR~-5 indicates that choosing a word over its synonym can introduce significantly larger impact on noise-induced comprehension errors in noisy environments. 
At SNR 0, the average HRS of the more intelligible synonym in the pair was 0.97, while the average HRS of the harder to understand synonym was 0.69; at SNR -5, the harder to comprehend synonym had an average HRS of 0.37, while the more intelligible one in the pair had an average HRS of 0.77.
These results mean that the relative gain in comprehension by choosing a lexical unit over its synonym which is high risky to be misheard in babble noise is 40\% at SNR~0 and 100\% at SNR~-5. Our experiment hence demonstrated that synonyms can differ substantially in intelligibility, especially at higher levels of noise. We therefore conclude that lexical replacement is a promising avenue for mitigating misperception in noise. The next step is to test whether this effect also holds for words in context.

\section{Synonyms \textit{with} context}
\label{sec:with_context}

A short utterance listening experiment was designed to study the recognition differences between synonym pair words in noisy environments, when the words were presented in linguistic context.
Participants of this listening experiment were asked to listen to noisy spoken words at three listening setups: babble noise at SNR 5, 0 and -5. 
Unlike the noise levels in the single word listening experiment, a low noise environment was considered instead of no noise as recognition in a quiet environment was close to a ceiling effect. 

\textbf{Stimuli:} For generating stimuli for this experiment, initially a list of top most 500 frequent words from the Spoken BNC2014 corpus \cite{love2017spoken} and their synonyms from WordNet\cite{wordnet} was created. Next, this list was filtered to identify synonym pairs such that both words from a pair semantically fit in a short utterance taken from the Spoken BNC2014 corpus. For example, for the synonym pair (\textit{sea}, \textit{ocean}), the following short utterances were used.

\begin{itemize}
\item \textit{and he runs away scared and dives into the \textbf{\underline{sea}}}
\item \textit{and he runs away scared and dives into the \textbf{\underline{ ocean}}}
\end{itemize}

This procedure resulted in 91 paired paraphrases, which were synthesized using gTTS. Subsequently, babble noise from NOISEX-92 \cite{varga1993assessment} was added. For the noise mixing, the SNR was kept fixed across the target as well as its context.

\textbf{Design and Procedure:} 
 We used the participants' transcription of what they hear to identify whether words were recognized correctly. Since the position of synonyms was not fixed across all utterances, participants were instructed to transcribe the whole utterance.
To mark those words which they couldn't perceive in an utterance, they were informed to use '...' (3 dots) as a placeholder.
Every stimulus was presented to six different participants in such a way that synonyms were not presented to the same participant.

\textbf{Participants:} A total of 51 native British English speakers (36 females and 15 males) with an average age of 34 (ranges from 20 to 50) participated in this experiment.

\textbf{Analysis:} Participants' responses were processed to identify whether target words (i.e.\ synonyms which undergo the lexical replacement in pair paraphrases) were recognized or not. For each stimulus, we again calculated HRS (as defined in (\ref{eq:HRS})). 
Further diff.HRS (as defined in (\ref{eq:diff.HRS})) was calculated for each paraphrase pair under different listening environments.
From this experiment, we expect to find that synonyms' recognition would be significantly different, even when they were presented with linguistic context in noise environments.

\subsection{Results}

\begin{figure}
    \centering
    \includegraphics[width=0.8\linewidth]{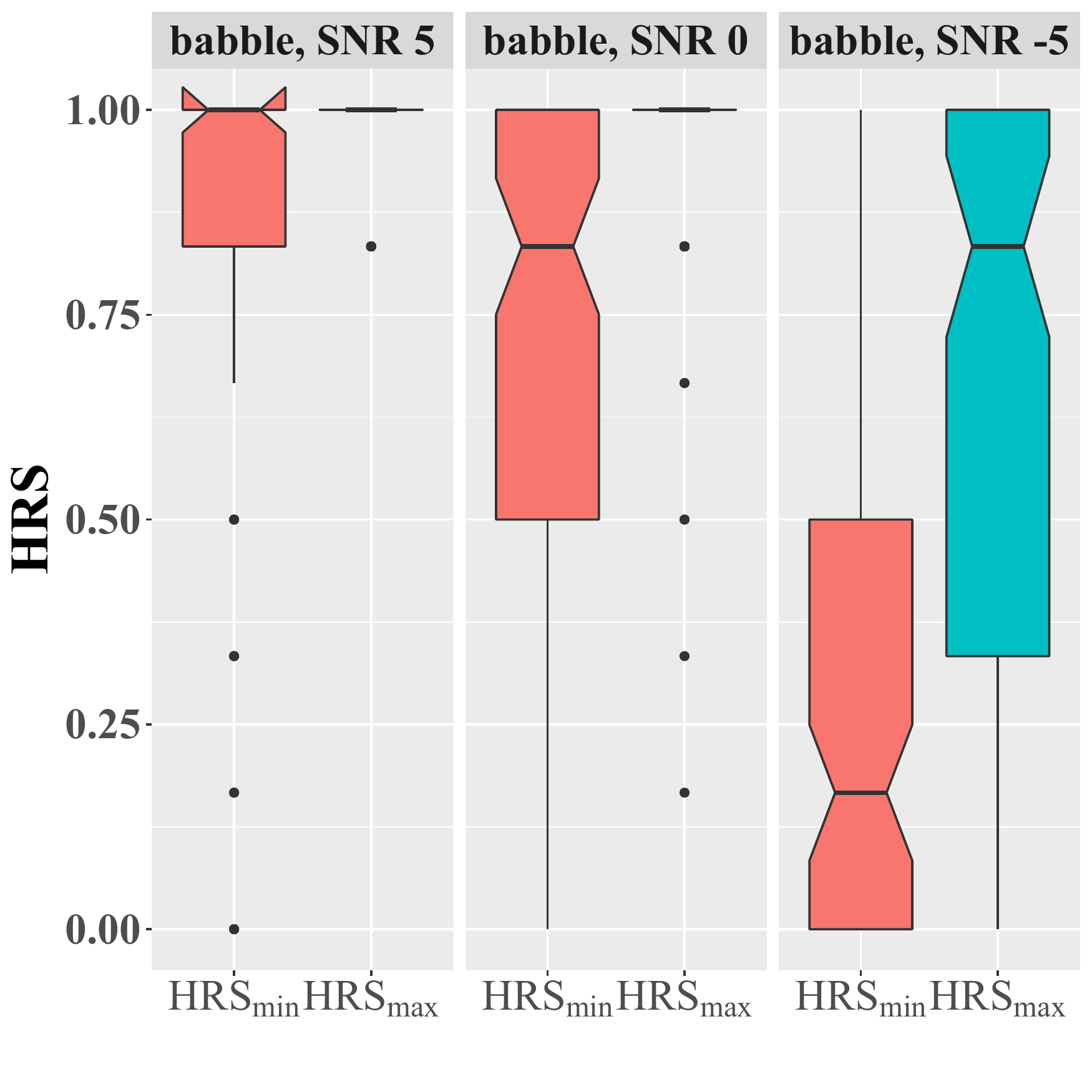}
    \caption{Distinction between less intelligible and more intelligible synonyms, when they were presented \textbf{with context} under different listening environments.}
    \label{fig:SUL_diffHRS_boxplot}
\end{figure}

After identifying the more intelligible synonym in a pair based on the HRSs, we can again compare the recognition scores in different noise types, and estimate the effect size of a synonym replacement strategy in noisy listening. 
$HRS_{min}$ and $HRS_{max}$ are used to refer to the HRS of the less intelligible and more intelligible synonyms in a pair.
Figure~\ref{fig:SUL_diffHRS_boxplot} summarizes the intelligibility differences of synonyms when they were presented with linguistic context in noisy environments.
The effect of replacing a target word with its synonym is evidently largest for high noisy environments. The mean difference in recognition
score between a target and its synonym, at SNR~-5 (0.37, \textit{p} $<$ 0.001) is significantly higher than in SNR~5 (0.15). However, the observed average difference at SNR~0 (0.21, \textit{p} $=$ 0.10) is not significantly different from that in SNR~5.

This indicates that lexical replacement is most beneficial when there is a large amount of noise. 
The average $HRS_{min}$ (0.84) and average $HRS_{max}$ (0.98) of synonyms at SNR~5 highlights that most of the target words were correctly recognized and this limits the scope of improvement that can be achieved by lexical replacement.
At SNR~0, the average of $HRS_{min}$ and $HRS_{max}$ was 0.73 and 0.94 respectively. 
In contrast to these environments, the average of $HRS_{min}$ and $HRS_{max}$ at SNR~-5 was 0.29 and 0.66 and this assured that at high noisy environments, lexical replacement introduced significant reduction in noise-induced comprehension errors.
This observed distinction in synonyms' recognition, even when they were preceded by naturalistic context such as in everyday conversations \cite{love2017spoken} implicates the usefulness of this strategy for generating noise-robust utterances.


\section{Modeling recognition differences}
\label{sec:modeling}

The two experiments have proven that lexical replacement can be a promising approach for improving speech intelligibility in noisy environments. In order to automatically be able to choose the more intelligible synonym, it is necessary to classify word intelligibility automatically. In this section, we explore the extent to which computational measures can
explain the variance of word recognitions in noise.

Previous work on misperceptions of words in context provided evidence that predictability of the target item is a significant factor for its intelligibility in noise.
We evaluated the following acoustic and linguistic features of a target word in context, by fitting a linear regression model (using its implementation in R \cite{Rlanguage}) to the data of 3 different noisy listening environments, separately.

\textbf{(1) Linguistic predictability:}
A pre-trained LSTM-based language model\cite{merityRegOpt}  was utilized to determine the predictability of a target word considering its left context in an utterance. The language model was trained on the transcription of a spoken corpus\cite{10.5555/1895550.1895693}, since stimuli utterances were taken from a speech corpus too. Log probability retrieved from this language model (hereafter referred as \textit{log.prob}), is used to represent how predictable a target word given its linguistic context.

\textbf{(2) Length of phonetic transcription:} Number of phonemes is particularly an important lexical feature for word recognition, as several studies have shown that longer words (which have fewer neighbors \cite{pisoni1985speech}) are easier to recognize \cite{vitevitch2002naturalistic, vitevitch2005neighborhood}. To study their effect on word intelligibility in noise, the length of phonetic transcription (which was generated using a Grapheme-to-Phoneme(G2P) converter \cite{bigPhoney}) was used to represent this feature (hereafter referred as \textit{ph.len}).


\textbf{(3) STOI: }
Short-Time Objective Intelligibility (STOI) \cite{taal2010short} measure is one of the classical Speech Intelligibility (SI) metrics. By comparing temporal envelopes of clean and noisy speech it captures the mean correlation between the energy of clean and distorted time-frequency units over all frames and bands. STOI value ranges from -1.0 to 1.0, representing least intelligible to most intelligible audio signals. Three STOI values for each target word were then calculated by considering the clean and noisy signals from all 3 listening environments.

\subsection{Results on recognition}

As a first analysis, significance of above mentioned features for determining the HRS in noise was evaluated by fitting a linear regression model separately for each of the listening environment data. 
At SNR~5, the model identified log probability ($\hat{\beta}$ = 0.031, \textit{SE} = 0.01, \textit{t} = 2.16, \textit{p} $<$ 0.05) as the only significant feature for explaining the variance in human recognition. At SNR~0, both log predictability ($\hat{\beta}$ = 0.05, \textit{SE} = 0.02, \textit{t} = 2.66, \textit{p} $<$ 0.05) and phoneme length ($\hat{\beta}$ = 0.04, \textit{SE} = 0.02, \textit{t} = 2.14, \textit{p} $<$ 0.05) were significant predictors of HRS.

However, at babble SNR~-5, we find that phoneme length ($\hat{\beta}$ = 0.06, \textit{SE} = 0.03, \textit{t} = 2.09, \textit{p} $<$ 0.05) and \textit{STOI} ($\hat{\beta}$ = 0.06, \textit{SE} = 0.03, \textit{t} = 2.09, \textit{p} $<$ 0.05) are significant predictors of HRS, but not predictability.
This difference between noise conditions may be due to difficulty with decoding the context: if the context cannot be fully understood, then it cannot be used effectively for predicting upcoming words.

\subsection{Analysis of recognition differences between synonyms}
Next, we separate out overall effects of predictability of a word from the difference in predictability between the two synonyms, in order to not only observe whether predictability as such is a significant predictor of HRS but also whether the difference in predictability between the synonyms makes a difference. Therefore, we encoded the response variable in terms of the difference between the HRS scores in a pair of synonyms by subtracting the HRS of the less intelligible synonym from the HRS of the more intelligible synonym. The resulting \textit{diff.HRS} scores thus range between 0 and 1, with 0 indicating that there was no difference in intelligibility.

Furthermore, we used the variable \textit{log.prob} to encode the predictability of the more intelligible word in the pair, and separately encoded the difference between them in the variable \textit{diff.log.prob}. Positive values of \textit{diff.log.prob} thus mean that the synonym with higher HRS was also more predictable.
Similarly, we separately encoded the word length of the better recognized synonym in a pair as \textit{ph.len}, and the difference in length to the other synonym as \textit{diff.ph.len}.
In addition, for each listening environment the intelligibility measure based on the acoustic features of the most recognized synonym in a pair was encoded as \textit{STOI} and its difference with the other synonym as \textit{diff.STOI}.



\begin{table}[th]
    \caption{ Evaluation of HRS difference between synonyms by fitting linear regression models to short utterance listening experiment data of \textbf{babble noise at SNR~5, SNR~0 and SNR~-5}.}
    \label{tab:SWL_n1_n3_n5}
    \centering
    \begin{tabular} {llllll}
    \toprule
    & \textbf{$\hat{\beta}$} & \textbf{SE} & \textbf{t value} & \textbf{p-value}              \\  \midrule
    Babble, SNR 5 \\
    \bottomrule
    (Intercept)      & -0.775 & 0.501 & -1.546 & 0.126             \\
    \textbf{\textit{log.prob} }        & -0.034 & 0.014 & -2.477 & 0.015 \textbf{*}   \\
    \textbf{\textit{diff.log.prob}}    & 0.033 & 0.009 & 3.651 & 0.0 \textbf{***} \\
    \textbf{\textit{ph.len}}       & 0.027 & 0.014 & 1.991 & 0.050 \textbf{*}   \\
    \textbf{\textit{diff.ph.len}}  & 0.023 & 0.01 & 2.254 & 0.027 \textbf{*}   \\
    \textit{STOI}      & 0.464 & 0.507 & 0.915 & 0.363            \\
    \textit{diff.STOI}    & -0.664 & 0.386 & -1.722 & 0.089 .      \\
    \bottomrule
    \bottomrule
    Babble, SNR 0 \\
    \bottomrule
    (Intercept)      & -0.608 & 0.43 & -1.415 & 0.161                \\
    \textbf{\textit{log.prob} }        & -0.045 & 0.015 & -2.967 & 0.004 \textbf{**}  \\
    \textbf{\textit{diff.log.prob}}    & 0.04 & 0.011 & 3.595 & 0.001 \textbf{***} \\
    \textit{ph.len }      & 0.011 & 0.015 & 0.712 & 0.478               \\
    \textbf{\textit{diff.ph.len}}  & 0.033 & 0.012 & 2.756 & 0.007 \textbf{**}  \\
    \textit{STOI}         & 0.346 & 0.491 & 0.704 & 0.483               \\
    \textit{diff.STOI}    & -0.175 & 0.33 & -0.531 & 0.597   \\
    \bottomrule
    \bottomrule
    Babble, SNR -5 \\
    \bottomrule
    (Intercept)      & 1.134 & 0.449 & 2.523 & 0.014  *           \\
    \textit{log.prob}         & -0.018 & 0.019 & -0.927 & 0.356               \\
    \textit{diff.log.prob}    & 0.011 & 0.013 & 0.836 & 0.406              \\
    \textit{ph.len }      & -0.009 & 0.019 & -0.473 & 0.637              \\
    \textit{diff.ph.len}  & 0.025 & 0.015 & 1.749 & 0.084  .           \\
    \textbf{\textit{STOI} }        & -1.428 & 0.495 & -2.887 & 0.005 \textbf{**} \\
    \textbf{\textit{diff.STOI} }   & 0.694 & 0.324 & 2.142 & 0.035 \textbf{*}  \\
    \bottomrule
    \bottomrule
    \end{tabular}
    \end{table}
    
For the analysis, maximal models with all features were considered and the best fitting model (which has the lowest Akaike Information Criterion (AIC)) selection was performed using the step function in R \cite{Rlanguage}.
The maximal model at SNR~5 identified the difference in a synonyms' predictability in context as well as their difference in number of phonemes as significant features in explaining the variance in diff.HRS; \textit{diff.log.prob} ($\hat{\beta}$ = 0.03, \textit{SE} = 0.01, \textit{p} $<$ 0.001) and  \textit{diff.ph.len} ($\hat{\beta}$ = 0.02, \textit{SE} = 0.01, \textit{p} $<$ 0.05), see also Table~\ref{tab:SWL_n1_n3_n5}.
As these predictors are in the direction of the response variable, it indicates that replacing a lexical unit with its synonym which has better predictability in a context leads to better recognition under a noisy environment in which the context is intelligible. 
Similarly, the model shows that there is a gain in recognition when a lexical unit is replaced by its synonym which has more number of phonemes. 
These observations are congruent with earlier studies on the effect of predictability on word recognition \cite{kalikow1977development} and the reduction of confusions with longer words \cite{vitevitch2002naturalistic, vitevitch2005neighborhood}.
It is noteworthy that the difference in STOI was not significant in the maximal model and hence the best fit model excluded acoustic based features for explaining the variance in diff.HRS at low noisy environment.

The model exhibited similar effects at medium noisy environment (SNR~0) by identifying the significance of  \textit{diff.log.prob} ($\hat{\beta}$ = 0.04, \textit{SE} = 0.01,  \textit{p} $<=$ 0.001) and \textit{diff.ph.len} ($\hat{\beta}$ = 0.03, \textit{SE} = 0.01,  \textit{p} $<$ 0.01) for explaining the variance in improved recognition. 
This reflects that under low/medium noisy environments, the gain in recognition introduced by lexical replacement is better explained by the improved predictability or the increased number of phonemes introduced by the replaced lexical item.
However, the difference between the intelligibility of synonyms didn't have an effect on their recognition in these noisy environments.

In contrast, the model for SNR~-5 showed that neither \textit{diff.log.prob} nor \textit{diff.ph.len} were significant predictors of the improvement in HRS through lexical replacement.
Instead, it revealed that replacement of a lexical unit with its more intelligible synonym can be predicted by the measure $STOI$ and \textit{diff.STOI} ($\hat{\beta}$ = 0.69, \textit{SE} = 0.32,  \textit{p}  $<$ 0.05). This reflects that at high noisy environments, choosing a lexical unit which has better noise-robust acoustic cues than its synonym can significantly improve its recognition.







\section{Conclusion}
\label{sec:conclusion}

In this current paper, we studied the potential of a new strategy of choosing noise-robust lexical paraphrases to mitigate comprehension errors which are caused by noisy listening environments.
Listening experiments were conducted to investigate whether the recognition of synonyms differs in noisy environments and we found that the potential impact of lexical replacement increased with an increase in noise level. 
Similar effects were observed also when synonyms were presented with linguistic context in noise.

Further investigation on this reduction in noise-induced comprehension errors by lexical replacement revealed that the intelligibility of a word in low and medium noise conditions is primarily driven by a word's predictability.
On the other hand, in more noisy environments, the intelligibility of a word was mainly driven by its acoustic features as captured by the STOI.
Thus, our results highlight that when an intended meaning needs to be realized as spoken words in very noisy environments, choosing noise-robust lexical paraphrases is a promising approach to improve comprehension.
We believe the current study can contribute in building better solutions to address hearing difficulties of individuals in adverse listening environments.

\section{Acknowledgements}

Funded by the DeutscheForschungsgemeinschaft (DFG, German Research Foundation) – Project-ID 232722074 – SFB 1102.


\begin{thebibliography}{10}
\providecommand{\url}[1]{#1}
\csname url@samestyle\endcsname
\providecommand{\newblock}{\relax}
\providecommand{\bibinfo}[2]{#2}
\providecommand{\BIBentrySTDinterwordspacing}{\spaceskip=0pt\relax}
\providecommand{\BIBentryALTinterwordstretchfactor}{4}
\providecommand{\BIBentryALTinterwordspacing}{\spaceskip=\fontdimen2\font plus
\BIBentryALTinterwordstretchfactor\fontdimen3\font minus
  \fontdimen4\font\relax}
\providecommand{\BIBforeignlanguage}[2]{{%
\expandafter\ifx\csname l@#1\endcsname\relax
\typeout{** WARNING: IEEEtran.bst: No hyphenation pattern has been}%
\typeout{** loaded for the language `#1'. Using the pattern for}%
\typeout{** the default language instead.}%
\else
\language=\csname l@#1\endcsname
\fi
#2}}
\providecommand{\BIBdecl}{\relax}
\BIBdecl

\bibitem{grimshaw1980mishearings}
A.~D. Grimshaw, ``Mishearings, misunderstandings, and other nonsuccesses in
  talk: A plea for redress of speaker-oriented bias,'' \emph{Sociological
  inquiry}, vol.~50, no. 3-4, pp. 31--74, 1980.

\bibitem{brumm2011evolution}
H.~Brumm and S.~A. Zollinger, ``The evolution of the lombard effect: 100 years
  of psychoacoustic research,'' \emph{Behaviour}, vol. 148, no. 11-13, pp.
  1173--1198, 2011.

\bibitem{zollinger2011lombard}
S.~A. Zollinger and H.~Brumm, ``The lombard effect,'' \emph{Current Biology},
  vol.~21, no.~16, pp. R614--R615, 2011.

\bibitem{patel2008influence}
R.~Patel and K.~W. Schell, ``The influence of linguistic content on the lombard
  effect,'' \emph{Journal of Speech, Language, and Hearing Research}, 2008.

\bibitem{valentini2014using}
C.~Valentini-Botinhao and M.~Wester, ``Using linguistic predictability and the
  lombard effect to increase the intelligibility of synthetic speech in
  noise,'' in \emph{Fifteenth Annual Conference of the International Speech
  Communication Association}, 2014.

\bibitem{kalikow1977development}
D.~N. Kalikow, K.~N. Stevens, and L.~L. Elliott, ``Development of a test of
  speech intelligibility in noise using sentence materials with controlled word
  predictability,'' \emph{The Journal of the Acoustical Society of America},
  vol.~61, no.~5, pp. 1337--1351, 1977.

\bibitem{uslar2011does}
V.~Uslar, E.~Ruigendijk, C.~Hamann, T.~Brand, and B.~Kollmeier, ``How does
  linguistic complexity influence intelligibility in a german audiometric
  sentence intelligibility test?'' \emph{International Journal of Audiology},
  vol.~50, no.~9, pp. 621--631, 2011.

\bibitem{carroll2013effects}
R.~Carroll and E.~Ruigendijk, ``The effects of syntactic complexity on
  processing sentences in noise,'' \emph{Journal of psycholinguistic research},
  vol.~42, no.~2, pp. 139--159, 2013.

\bibitem{van2018speech}
E.~C. van Knijff, M.~Coene, and P.~J. Govaerts, ``Speech understanding in noise
  in elderly adults: the effect of inhibitory control and syntactic
  complexity,'' \emph{International journal of language \& communication
  disorders}, vol.~53, no.~3, pp. 628--642, 2018.

\bibitem{luce1998recognizing}
P.~A. Luce and D.~B. Pisoni, ``Recognizing spoken words: The neighborhood
  activation model,'' \emph{Ear and hearing}, vol.~19, no.~1, p.~1, 1998.

\bibitem{mcardle2008predicting}
R.~McArdle and R.~H. Wilson, ``Predicting word-recognition performance in noise
  by young listeners with normal hearing using acoustic, phonetic, and lexical
  variables,'' \emph{Journal of the American Academy of Audiology}, vol.~19,
  no.~6, pp. 507--518, 2008.

\bibitem{doi:10.1121/1.1908983}
\BIBentryALTinterwordspacing
J.~M. Pickett, ``Perception of vowels heard in noises of various spectra,''
  \emph{The Journal of the Acoustical Society of America}, vol.~29, no.~5, pp.
  613--620, 1957. [Online]. Available: \url{https://doi.org/10.1121/1.1908983}
\BIBentrySTDinterwordspacing

\bibitem{doi:10.1121/1.1810292}
\BIBentryALTinterwordspacing
A.~Cutler, A.~Weber, R.~Smits, and N.~Cooper, ``Patterns of english phoneme
  confusions by native and non-native listeners,'' \emph{The Journal of the
  Acoustical Society of America}, vol. 116, no.~6, pp. 3668--3678, 2004.
  [Online]. Available: \url{https://doi.org/10.1121/1.1810292}
\BIBentrySTDinterwordspacing

\bibitem{weber2003consonant}
A.~Weber and R.~Smits, ``Consonant and vowel confusion patterns by american
  english listeners,'' in \emph{15th International Congress of Phonetic
  Sciences [ICPhS 2003]}, 2003.

\bibitem{doi:10.1121/1.3224721}
\BIBentryALTinterwordspacing
T.~Jürgens and T.~Brand, ``Microscopic prediction of speech recognition for
  listeners with normal hearing in noise using an auditory model,'' \emph{The
  Journal of the Acoustical Society of America}, vol. 126, no.~5, pp.
  2635--2648, 2009. [Online]. Available:
  \url{https://doi.org/10.1121/1.3224721}
\BIBentrySTDinterwordspacing

\bibitem{lexicalneighborhoods}
\BIBentryALTinterwordspacing
C.~G. Clopper, J.~B. Pierrehumbert, and T.~N. Tamati, ``Lexical neighborhoods
  and phonological confusability in cross-dialect word recognition in noise,''
  \emph{Laboratory Phonology}, vol.~1, no.~1, pp. 65 -- 92, 2010. [Online].
  Available:
  \url{https://www.degruyter.com/view/journals/labphon/1/1/article-p65.xml}
\BIBentrySTDinterwordspacing

\bibitem{wilson2008comparison}
R.~H. Wilson and W.~B. Cates, ``A comparison of two word-recognition tasks in
  multitalker babble: Speech recognition in noise test (sprint) and
  words-in-noise test (win),'' \emph{Journal of the American Academy of
  Audiology}, vol.~19, no.~7, pp. 548--556, 2008.

\bibitem{doi:10.1121/1.4809540}
\BIBentryALTinterwordspacing
R.~Albert~Felty, A.~Buchwald, T.~M. Gruenenfelder, and D.~B. Pisoni,
  ``Misperceptions of spoken words: Data from a random sample of american
  english words,'' \emph{The Journal of the Acoustical Society of America},
  vol. 134, no.~1, pp. 572--585, 2013. [Online]. Available:
  \url{https://doi.org/10.1121/1.4809540}
\BIBentrySTDinterwordspacing

\bibitem{doi:10.1121/1.5090196}
\BIBentryALTinterwordspacing
M.~Cooke, M.~L. García~Lecumberri, J.~Barker, and R.~Marxer, ``Lexical
  frequency effects in english and spanish word misperceptions,'' \emph{The
  Journal of the Acoustical Society of America}, vol. 145, no.~2, pp.
  EL136--EL141, 2019. [Online]. Available:
  \url{https://doi.org/10.1121/1.5090196}
\BIBentrySTDinterwordspacing

\bibitem{marxer2016corpus}
R.~Marxer, J.~Barker, M.~Cooke, and M.~L. Garcia~Lecumberri, ``A corpus of
  noise-induced word misperceptions for english,'' \emph{The Journal of the
  Acoustical Society of America}, vol. 140, no.~5, pp. EL458--EL463, 2016.

\bibitem{cooke2009discovering}
M.~Cooke, ``Discovering consistent word confusions in noise,'' in \emph{Tenth
  Annual Conference of the International Speech Communication Association},
  2009.

\bibitem{vitevitch2002naturalistic}
M.~S. Vitevitch, ``Naturalistic and experimental analyses of word frequency and
  neighborhood density effects in slips of the ear,'' \emph{Language and
  speech}, vol.~45, no.~4, pp. 407--434, 2002.

\bibitem{verbmobil}
W.~Wahlster, ``Verbmobil,'' in \emph{Grundlagen und anwendungen der
  k{\"u}nstlichen intelligenz}.\hskip 1em plus 0.5em minus 0.4em\relax
  Springer, 1993, pp. 393--402.

\bibitem{wordnet}
C.~Fellbaum, \emph{WordNet: An Electronic Lexical Database}.\hskip 1em plus
  0.5em minus 0.4em\relax Bradford Books, 1998.

\bibitem{gTTS}
P.~N. Durette, \emph{Google Translate's text-to-speech API}, 2014 (accessed
  July 30, 2020), \url{https://pypi.org/project/gTTS/}.

\bibitem{varga1993assessment}
A.~Varga and H.~J. Steeneken, ``Assessment for automatic speech recognition:
  Ii. noisex-92: A database and an experiment to study the effect of additive
  noise on speech recognition systems,'' \emph{Speech communication}, vol.~12,
  no.~3, pp. 247--251, 1993.

\bibitem{palan2018prolific}
S.~Palan and C.~Schitter, ``Prolific. ac—a subject pool for online
  experiments,'' \emph{Journal of Behavioral and Experimental Finance},
  vol.~17, pp. 22--27, 2018.

\bibitem{love2017spoken}
R.~Love, C.~Dembry, A.~Hardie, V.~Brezina, and T.~McEnery, ``The spoken
  bnc2014: Designing and building a spoken corpus of everyday conversations,''
  \emph{International Journal of Corpus Linguistics}, vol.~22, no.~3, pp.
  319--344, 2017.

\bibitem{Rlanguage}
\BIBentryALTinterwordspacing
{R Core Team}, \emph{R: A Language and Environment for Statistical Computing},
  R Foundation for Statistical Computing, Vienna, Austria, 2019. [Online].
  Available: \url{https://www.R-project.org/}
\BIBentrySTDinterwordspacing

\bibitem{merityRegOpt}
S.~Merity, N.~S. Keskar, and R.~Socher, ``{Regularizing and Optimizing LSTM
  Language Models},'' \emph{arXiv preprint arXiv:1708.02182}, 2017.

\bibitem{10.5555/1895550.1895693}
J.~J. Godfrey, E.~C. Holliman, and J.~McDaniel, ``Switchboard: Telephone speech
  corpus for research and development,'' in \emph{Proceedings of the 1992 IEEE
  International Conference on Acoustics, Speech and Signal Processing - Volume
  1}, ser. ICASSP'92.\hskip 1em plus 0.5em minus 0.4em\relax USA: IEEE Computer
  Society, 1992, p. 517–520.

\bibitem{pisoni1985speech}
D.~B. Pisoni, H.~C. Nusbaum, P.~A. Luce, and L.~M. Slowiaczek, ``Speech
  perception, word recognition and the structure of the lexicon,'' \emph{Speech
  communication}, vol.~4, no. 1-3, pp. 75--95, 1985.

\bibitem{vitevitch2005neighborhood}
M.~S. Vitevitch and E.~Rodr{\'\i}guez, ``Neighborhood density effects in spoken
  word recognition in spanish,'' \emph{Journal of Multilingual Communication
  Disorders}, vol.~3, no.~1, pp. 64--73, 2005.

\bibitem{bigPhoney}
R.~Epp, \emph{BigPhoney, a python module}, 2018 (accessed by July 30, 2020),
  \url{https://github.com/repp/big-phoney}.

\bibitem{taal2010short}
C.~H. Taal, R.~C. Hendriks, R.~Heusdens, and J.~Jensen, ``A short-time
  objective intelligibility measure for time-frequency weighted noisy speech,''
  in \emph{2010 IEEE international conference on acoustics, speech and signal
  processing}.\hskip 1em plus 0.5em minus 0.4em\relax IEEE, 2010, pp.
  4214--4217.

\end{thebibliography}
\end{document}